\documentclass[11pt]{article}
\usepackage{acl2015}
\usepackage{times}
\usepackage{url}
\usepackage{latexsym}
\usepackage{graphicx} 
\usepackage{booktabs}
\usepackage{hyperref}
\usepackage{natbib}
\title{I or Not I: Unraveling the Linguistic Echoes of Identity in Samuel Beckett's "Not I" Through Natural Language Processing}

\author{Arezou Zahiri Pourzarandi \\
  {\tt a.zahiri@student.art.ac.ir} \\\And
  Farshad Jafari \\
  {\tt farshad.jafari@gatech.edu} \\}

\date{December 2023}

\begin{document}
\maketitle

\begin{abstract}
Exploring the depths of Samuel Beckett's "Not I" through advanced natural language processing techniques, this research uncovers the intricate linguistic structures that underpin the text. By analyzing word frequency, detecting emotional sentiments with a BERT-based model, and examining repetitive motifs, we unveil how Beckett's minimalist yet complex language reflects the protagonist's fragmented psyche. Our results demonstrate that recurring themes of time, memory, and existential angst are artfully woven through recursive linguistic patterns and rhythmic repetition. This innovative approach not only deepens our understanding of Beckett's stylistic contributions but also highlights his unique role in modern literature, where language transcends simple communication to explore profound existential questions.
\end{abstract}

\section{Introduction}

This paper explores Samuel Beckett's play "Not I," focusing on its unique narrative structure and profound thematic depth. "Not I" is characterized by its minimalistic approach, where the entirety of the play revolves around a single character, known as "Mouth," who delivers a rapid, fragmented monologue that plunges the audience into the depths of existential despair and identity crisis. Through this study, we aim to analyze the linguistic intricacies and the existential themes that dominate the play, offering insights into Beckett's use of language as a tool for expressing the human condition.

Our analysis is particularly interested in the play's exploration of themes such as isolation, identity, and the struggle for meaning within the framework of existential philosophy. The monologue's repetitive and disjointed nature not only reflects the protagonist’s fragmented psyche but also highlights Beckett's stylistic departure from traditional narrative forms, emphasizing a stream-of-consciousness technique that challenges the boundaries of theatrical presentation.

This paper seeks to dissect the layers of complexity in Beckett's text, drawing connections between the play's stylistic elements and its broader existential themes. 
In exploring the use of minimalism in "Not I" and its ability to evoke profound emotional responses, we hope to enhance the understanding of its role within Beckett's body of work and the broader context of modern theatre.

\subsection{Beckett's writing style}

Samuel Beckett's writing style is characterized by minimalism and the economical use of language. He employs sparse dialogue and simple vocabulary, focusing on the essence of human experience. Repetition of words, phrases, and scenes is a hallmark of his work, emphasizing the monotony and cyclical nature of existence. Beckett explores existential themes such as absurdity, isolation, and despair, often using non-linear narratives and long introspective monologues. His settings are minimalist, with rich symbolism and imagery reflecting the emptiness of the human condition. Beckett's innovative use of language and structure challenges traditional theatrical norms.

\subsection{NOT I}
"Not I" by Samuel Beckett is a dramatic monologue characterized by a single character, "Mouth," who delivers a rapid and fragmented stream of consciousness. The narrative is chaotic and disjointed, reflecting the character's inner turmoil and fragmented mental state. Beckett uses repetitive language and simple, stark vocabulary to create a rhythmic, almost musical quality. The play explores themes of existential despair, identity, isolation, and trauma, with the disembodied mouth symbolizing the compulsion to speak despite the futility of communication. Influenced by existential philosophy, "Not I" delves into the absurdity of human existence and the search for meaning in a seemingly meaningless world.

\subsection{Objectives}
To elucidate the complex layers of text within "Not I", we conducted three distinct analyses:

\begin{itemize}

\item\textbf{{Word Frequency Analysis}}
We utilized a word cloud to visually represent the most frequently occurring words within the text, providing insights into the thematic and narrative emphasis of the play

\item \textbf{{Emotional Analysis}}
Employing a BERT-based transformer model for emotion recognition, we analyzed the emotional undertones conveyed through the monologue, charting the prevalence and shifts in emotional intensity throughout the play.

\item \textbf{{Repetitive Pattern Analysis}}
Through two specific plots, we examined the occurrences of repetitive words and the longest repetitive motives, shedding light on the play's structure and Beckett's use of repetition to emphasize the cyclical nature of the protagonist’s thoughts.
\end{itemize}

\section{Background}

The analysis of rhythm and repetition in literary texts has long been a focus of both theoretical and computational research. Early approaches to rhythm in literature often centered on rhetorical structures and prosody. For instance, studies have explored the role of rhythm as a communicative device, using linguistic structures such as syllabic patterns and punctuation to establish rhythmic units within texts \citep{balint2016rhetorical}. Computational methods have since evolved, automating the detection of rhythmic figures, which has proven valuable in analyzing prose across centuries \citep{lagutina2020automatic}. However, these methods, though useful for identifying surface-level rhythmic patterns, often lack the sophistication to handle more complex, non-linear narratives like Samuel Beckett's \textit{Not I}. Most rhythm analysis tools, particularly in earlier studies, have been designed for languages or texts with predictable prosodic patterns, such as French poetry \citep{boychuk2014automated}, leaving a gap in handling experimental works that rely on disjointed, rapid speech.

Repetition is another critical element that shapes both form and meaning in literary texts. \citep{tannen2007talking} highlights how repetition in dialogue establishes coherence and amplifies emotional tones, a principle particularly relevant to dramatic works. Studies have classified repetition into various forms—such as positional, associative, and aggregative—focusing on how these forms shape textual structures \citep{altmann2015forms}. \textbf{Positional repetition} refers to the recurrence of units in specific locations within the text, such as at the beginning or end of sentences, which can create patterns of emphasis and rhythm. \textbf{Associative repetition} occurs when words or phrases are repeated alongside certain other words, suggesting deeper semantic links. \textbf{Aggregative repetition} involves clusters of repeated units appearing in concentrated sections of a text, often creating a sense of intensity or focus around particular ideas or emotions. These different forms of repetition are not only stylistic choices but also serve to reinforce themes and motifs within a text. Despite the substantial insights gained from these classifications, their application to more fragmented and non-linear texts, such as Beckett’s \textit{Not I}, has not been thoroughly explored, particularly in how repetition might reflect a disordered mental state or fragmented identity.

Current approaches to sentiment and emotion analysis in literary texts have predominantly focused on character-to-character interactions in works like Shakespeare’s plays, where sentiment analysis tools are used to track emotional shifts through dialogue \citep{nalisnick2013character}. These methods, however, often assume more stable narratives, whereas Beckett’s work presents emotional fragmentation. Emotion analysis of literature has expanded into generating high-quality emotion arcs, offering a more dynamic view of character emotions \citep{teodorescu2022frustratingly}, but these approaches have yet to be applied to monologues that, like \textit{Not I,} intentionally disrupt narrative flow to reflect fractured identity. Our research addresses this by leveraging today's deep learning models, which allow for more sophisticated tracking of emotional shifts in fragmented, non-linear speech patterns, exceeding the capacity of earlier lexicon-based tools \citep{yavuz2020analyses}.

Additionally, discourse analysis tools have traditionally focused on the logical structure of dialogue and narrative \citep{Crossley2014AnalyzingDP}, but these have not been extensively adapted to the extreme minimalism found in Beckett’s monologue. The absence of clear interlocutors and the focus on internalized speech require a more advanced level of analysis. Recent advancements in Natural Language Processing (NLP), such as transformer-based models, allow for a deeper and more nuanced exploration of these texts \citep{tunstall2022natural}, enabling us to model the rhythmic fragmentation and its psychological impact.

Existing rhythm detection tools provide valuable insights into metrical structures and patterns, yet they remain limited in their application to texts that break conventional rhythmic forms, such as Beckett’s use of rapid, disjointed speech. Earlier rhythm extraction models rely on pre-defined patterns and lack the flexibility to adapt to experimental literature \citep{niculescu2016rhythm}. We address this by integrating deep learning models that can analyze irregular, non-linear text structures, providing a more accurate depiction of the chaotic rhythm that defines \textit{Not I}.

Our research fills these gaps by applying advanced NLP techniques and deep learning models to analyze repetition, rhythm, and emotional fragmentation in \textit{Not I}. While earlier studies laid the groundwork for understanding literary rhythm and emotion, they have not adequately addressed works that defy conventional narrative structures. By using transformer models and state-of-the-art NLP tools, we are able to capture the complex interplay of rhythm, repetition, and identity in Beckett's monologue, offering a new framework for analyzing experimental literature that challenges both linguistic and narrative norms.

\section{Methodology}
To prepare the text of "Not I" for analysis, we implemented a systematic approach to ensure the integrity and quality of the data suitable for NLP. This approach involved extracting text, cleaning it, segmenting it, and using various NLP tools and techniques to carry out a detailed linguistic and emotional analysis. The codes used for this analysis are all available at the \href{https://github.com/frshdjfry/NLP-in-Dramatic-Literature}{GitHub repository}.

\subsection{Text Extraction and Preprocessing}
Text Extraction: We employed the \href{https://github.com/tesseract-ocr/tesseract}{Tesseract} Optical Character Recognition (OCR) tool to digitize the printed pages of "Not I," converting them into machine-readable text.
Cleaning: The digitized texts were meticulously cleaned to remove OCR errors, irrelevant formatting, and non-textual elements, ensuring the data's purity for subsequent analysis.
Text Segmentation: Recognizing the theatrical nature of the text, we segmented the play into 30-word units, each approximating 10 to 15 seconds of stage time, to analyze linguistic and emotional content in manageable, performance-relevant chunks.

\subsection{Linguistic Analysis}
Using the \href{https://spacy.io/}{spaCy} library configured for the English language, we conducted a detailed linguistic analysis, which included:

Word Frequency and Word Clouds: We identified the most common words and visualized them using WordCloud in \href{https://matplotlib.org/}{matplotlib}, highlighting key thematic elements of the text.
Vocabulary Richness: We computed lexical diversity to assess the linguistic complexity and stylistic variance across the play.

\subsection{Emotion Detection}
To capture a nuanced spectrum of emotions, we utilized a BERT-based transformer model via \href{https://huggingface.co/}{Hugging Face}'s pipeline for text classification \citep{hartmann2022emotionenglish}. This model provided scores for multiple emotions, enabling us to develop a detailed emotional profile of each segment.

\subsection{Repetitive Pattern Analysis}
We designed a Python script employing regular expressions and array manipulations to locate and document the positions of frequently occurring words. These positions were then visually represented through scatter plots on a Gantt chart, enhancing our understanding of word distribution and frequency across the text, which aided in identifying significant linguistic patterns.

\section{Results}
\label{sec}

\begin{figure*}[!htb]
    \centering
    \includegraphics[width=0.8\textwidth]{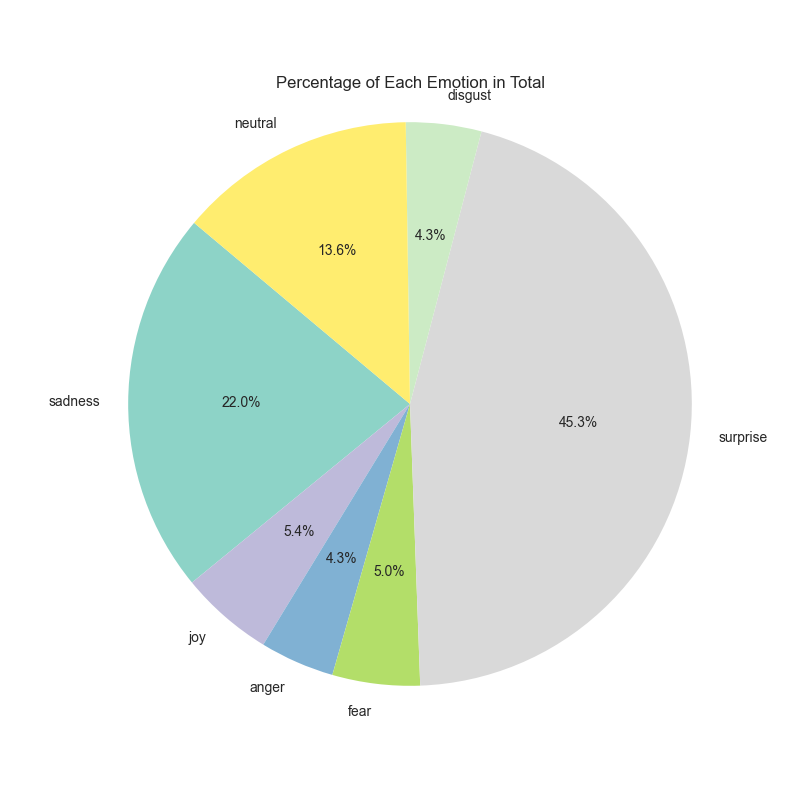} % Adjust the width to fit within the text more snugly
    \caption{Pie chart showing the distribution of emotions in "Not I."}
    \label{fig:emotion_pie_chart}
\end{figure*}

This section presents the analysis of the emotional landscape in Samuel Beckett's "Not I" as depicted through a pie chart of emotion distribution. Figure \ref{fig:emotion_pie_chart} illustrates the proportion of each emotion captured through the text analysis.

\subsection{Analysis of Emotions}

\subsubsection{Surprise}
The predominant emotion in "Not I" is \textit{surprise}, which is reflected in several aspects of the text:
\begin{itemize}
    \item \textbf{Linguistic Patterns}: The frequent usage of the exclamation "oh" highlights a continuous sense of astonishment.
    \item \textbf{Character Behavior}: The character "Mouth" consistently uses interrogatives like "when," "who," and "what," underscoring a state of bewilderment due to the lack of definitive answers.
    \item \textbf{Punctuation}: The absence of periods and the reliance on exclamation and question marks amplify the theme of surprise, with the text concluding with a dash ("-"), reinforcing the unresolved, astonished tone.
\end{itemize}

\subsubsection{Sadness}
\textit{Sadness} emerges as the second most prevalent emotion, supporting Beckett’s exploration of despair:
\begin{itemize}
    \item \textbf{Contextual Expectation}: Beckett's works are known for their bleak outlook, making the pervasive sadness in "Not I" a direct reflection of his narrative style.
    \item \textbf{Character Expression}: The fragmented and disjointed speech of "Mouth" conveys deep loss and hopelessness, mirroring the underlying sadness of the narrative.
\end{itemize}

\subsubsection{Neutrality}
\textit{Neutrality} is identified as the third significant emotional tone, intriguing given the play's minimalist approach:
\begin{itemize}
    \item \textbf{Language Use}: The rapid succession of simple, emotionally neutral words contributes to a dense, minimally expressive linguistic texture.
    \item \textbf{Narrative Style}: The disorganized delivery by "Mouth" suggests a struggle with disjointed memories, lending a neutral tone to the narrative.
    \item \textbf{Character's Condition}: The portrayal of "Mouth" as a character grappling with Alzheimer’s disease, attempting to piece together fragmented memories, reflects the simplicity and neutrality of the language, dominated by the necessity to speak.
\end{itemize}

\subsubsection{Other Emotions}
Other emotions present in the text exhibit a balanced distribution, consistent with the content of the play and contributing less critically to the overall emotional analysis compared to surprise, sadness, and neutrality.

\subsection{Quantitative Textual Analysis}
\label{sec:quantitative_textual_analysis}

\begin{figure*}[!htb]
    \centering
    \includegraphics[width=1.0\textwidth]{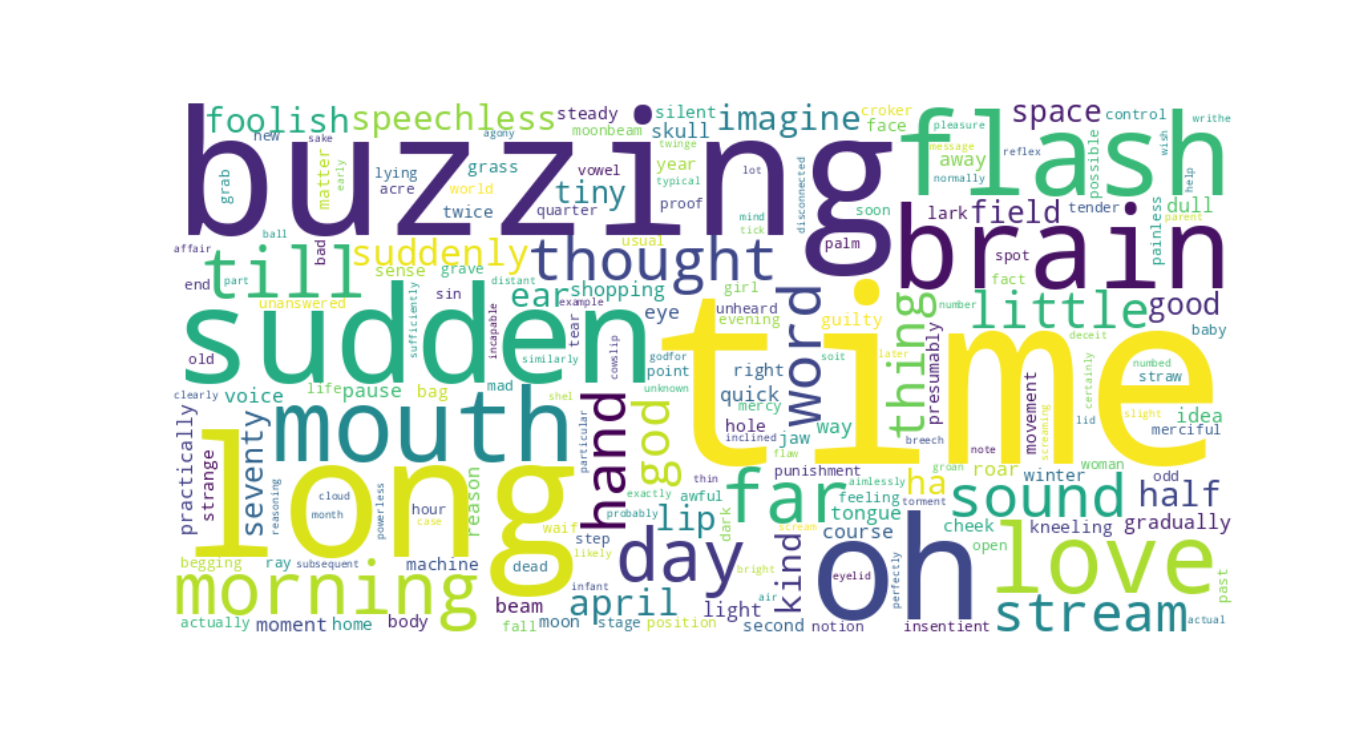}
    \caption{Word cloud depicting the predominant words in "Not I."}
    \label{fig:wordcloud}
\end{figure*}

In this section, we delve into the quantitative aspects of word usage in Samuel Beckett's "Not I," focusing on the most frequently occurring words as illustrated in Figure \ref{fig:wordcloud} and detailed in Table \ref{tab:word_frequency}.

\begin{table}[ht]
    \centering
    \begin{tabular}{lc}
    \hline
    Word & Frequency \\
    \hline
    Time & 17 \\
    Buzzing & 15 \\
    Long & 14 \\
    What? & 13 \\
    Like & 13 \\
    Oh & 10 \\
    Sudden & 10 \\
    \hline
    \end{tabular}
    \caption{Table of the top 7 most frequently occurring words in "Not I," excluding common stopwords.}
    \label{tab:word_frequency}
\end{table}

\subsubsection{Linguistic Significance of High-Frequency Words}
The analysis of high-frequency words provides insights into the thematic concerns and the narrative style employed by Beckett. The terms identified offer a window into the emotional and psychological state of the protagonist, "Mouth."

The frequent repetition of specific words is instrumental in underscoring the protagonist "Mouth's" psychological and emotional turmoil. Each key term contributes distinctively to the narrative's depth:

\begin{itemize}
    \item \textbf{"Time"} Everything in this play is about time. Specifically, it is about the passage of time, the deterioration of the mind due to the passage of time (aging), the erratic back and forth in time, the disjointed journey through time, the discovery of time, and ultimately the loss of time.
    \item \textbf{"Buzzing"} This repetition underscores the incessant nature of the sounds within the protagonist's skull, representing a relentless torment. When "Mouth" utters this word, it signifies moments where she momentarily takes control of the voice, screaming to identify the exact nature of these sounds. The "buzzing" symbolizes the overwhelming flood of fragmented and unclear memories that plague her mind.
    \item \textbf{"Long"} Despite everything being conveyed quickly in the text, the events are spaced far apart in the core of the story. the character stands at a great spatial and temporal distance from her memories at this moment and tries to search through these distances rapidly.
    \item \textbf{"Imagine"} The frequently repeated word "imagine" is used by "Mouth" to command and urge the brain to imagine and recall again. It is as if "Mouth", after saying and hearing a series of memories, suddenly realizes what was heard and wants to imagine it all over again.
    \item \textbf{"Sudden"} underlines the abrupt and often shocking re-emergence of repressed memories or thoughts.
\item Terms related to bodily parts such as \textbf{"Brain"}, \textbf{"Mouth"}, and \textbf{"Hand"} emphasize the physicality of the protagonist's existential and corporeal experience.
\end{itemize}

Through the meticulous repetition of these words, Beckett not only crafts a rhythmic and compelling narrative but also constructs a vivid portrayal of the protagonist’s fragmented psyche. The methodological approach combining computational linguistics and traditional literary analysis enhances our understanding of the play’s deep-seated themes of memory, identity, and existential angst.

\subsection{Repetitive Lexical Patterns}
\label{sec:repetitive_patterns}

\begin{figure*}[!htb]
    \centering
    \includegraphics[width=1.0\textwidth]{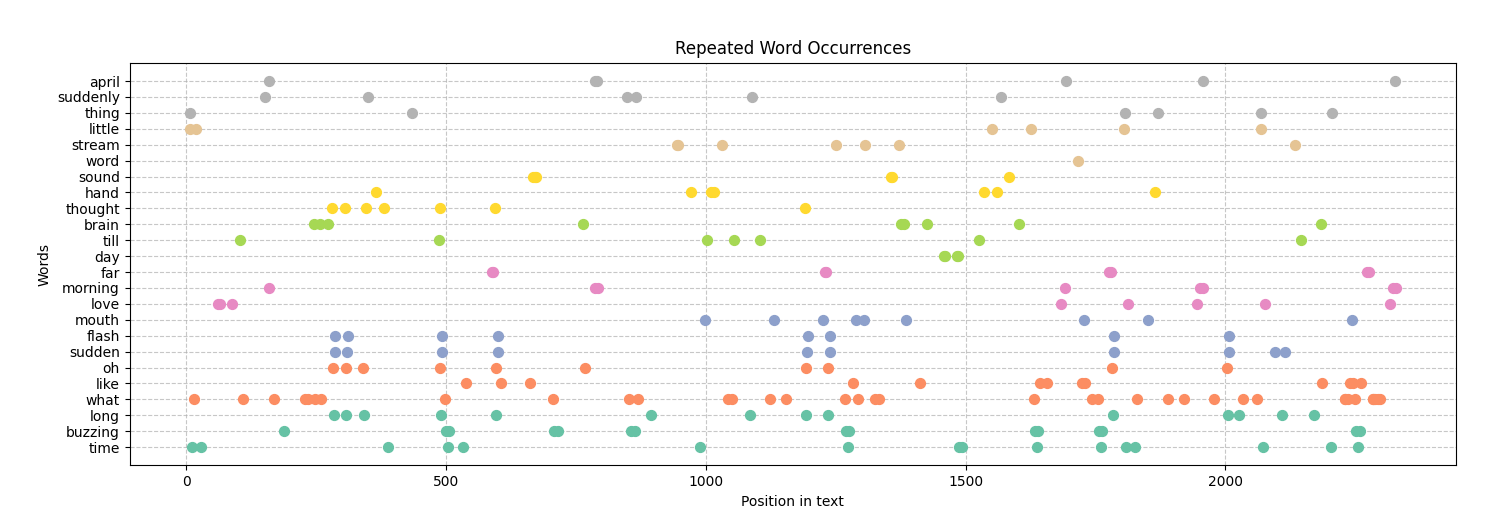}
    \caption{Gantt chart with scatter plot of word occurrences in "Not I," illustrating the clustering of lexical repetitions.}
    \label{fig:gantt_chart}
\end{figure*}

This study utilizes a Gantt chart with a scatter plot, as shown in Figure \ref{fig:gantt_chart}, to illustrate the distribution and clustering of word occurrences in Samuel Beckett's "Not I." The analysis highlights the implicit structuring of the text, which mirrors the traditional three-act format through its lexical repetitions.

\subsubsection{Implicit Structuring Through Lexical Clustering}
The text's division into three major clusters suggests an underlying three-act structure that Beckett does not explicitly denote. Each cluster corresponds to a distinct narrative phase within the play:

\begin{itemize}
    \item \textbf{Cluster One}: Encompassing the initial segments of the text, this cluster includes frequent repetitions of "buzzing," "time," and "long," which introduce the central themes and set the emotional tone.
    \item \textbf{Cluster Two}: Situated in the middle segments, this cluster features "mouth," "sudden," and "imagine," indicating a progression in the narrative and the protagonist's deepening engagement with her memories.
    \item \textbf{Cluster Three}: Found in the concluding segments, this includes repetitions of "oh," "what," and "far," reflecting the culmination of the narrative and the protagonist's unresolved queries.
\end{itemize}

\subsubsection{Linguistic and Narrative Insights}
The analysis of these clusters reveals not only the play's unique narrative structure but also Beckett's strategic use of language to deepen the narrative's complexity:
\begin{figure*}[!htb]
    \centering
    \includegraphics[width=1.0\textwidth]{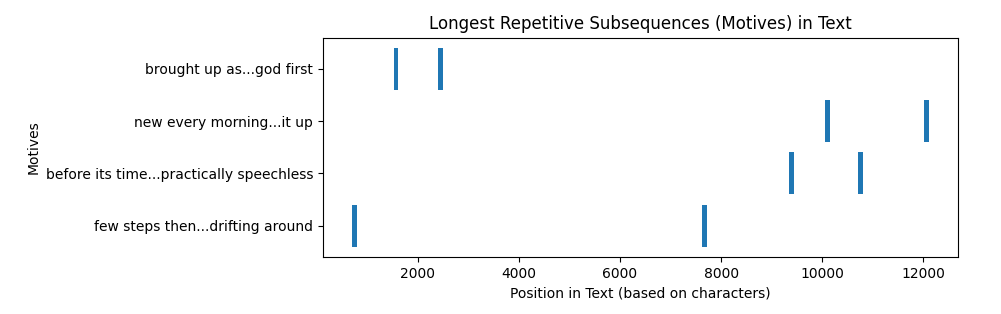}
    \caption{Chart depicting the longest repetitive subsequences in \textit{Not I}.}
    \label{fig:motif_chart}
\end{figure*}

\begin{itemize}
    \item \textbf{Rhythmic and Musical Language}: The repetitive nature of the selected words across clusters introduces a rhythmic, almost musical quality to the monologue, aiding in the portrayal of the protagonist's fluctuating psychological states.
    \item \textbf{Cognitive and Emotional Mapping}: The words within each cluster help map the cognitive and emotional journey of the character, "Mouth," illustrating her struggle with fragmented memories and existential questions.
\end{itemize}

\subsection{Longest Repetitive Subsequences (Motifs)}
\label{sec:motif_analysis}

This section discusses the analysis of the longest repetitive subsequences, or motifs, within the text of Samuel Beckett's \textit{Not I}. The motifs are crucial in understanding the rhythmic and structural composition of the text.

Figure~\ref{fig:motif_chart} shows a chart displaying the longest repetitive subsequences, with the first three words and the last two words of each repeated segment highlighted. Examples of such motifs include:
\begin{itemize}
    \item ``brought up as...god first''
    \item ``new every morning...it up''
    \item ``before its time...practically speechless''
    \item ``few steps then...drifting around''
\end{itemize}
These motifs recur at various points throughout the text, illustrating the structural repetition employed by Beckett.

\subsubsection{Significance of Motifs}
The motifs provide evidence that Beckett’s text employs paragraph-length structures in a rhythmic manner, akin to a musical composition. The entire text can be likened to a recursive pattern, where language is used to create a complex and layered auditory experience.

\subsubsection{Musical Structure in Text}
\begin{itemize}
    \item \textbf{Rhythmic Repetition}: The repetition of larger motifs creates a rhythmic flow within the narrative, mirroring the repetitive nature of individual words.
    \item \textbf{Recursive Patterns}: The text’s linguistic structure exhibits recursive patterns that enhance its musicality, reflecting the complexity and depth of the narrative.
    \item \textbf{Musical Analog}: This structural approach parallels the use of canons in music, where a melody is repeated and imitated at different intervals.
\end{itemize}

Through the meticulous repetition of these motifs, Beckett not only crafts a rhythmic and compelling narrative but also constructs a vivid portrayal of the protagonist’s fragmented psyche. The text’s recursive structure and its similarity to musical canons highlight the deliberate use of repetition to deepen the audience's engagement with the play’s themes and emotional landscape.
\section{Conclusion}
\label{sec}

This study has systematically dissected Samuel Beckett's "Not I" through advanced natural language processing techniques, revealing the profound complexity and structured repetition that underscore the play's essence. Our analyses have demonstrated that Beckett employs a meticulously recursive linguistic framework that mirrors the psychological fragmentation of the protagonist. The integration of word frequency analysis illuminated the thematic focus on time, memory, and existential turmoil, while emotional analysis underscored the predominant feelings of surprise and sadness, reflecting the play’s intense emotional landscape. Finally, the examination of repetitive motifs highlighted Beckett's use of language as a musical and fractal element, reinforcing the play’s rhythmic and recursive narrative structure. Collectively, these findings not only deepen our understanding of Beckett's stylistic  modulations but also underscore his unique contribution to modern literature, where language transcends simple communication to evoke profound existential reflections.

\bibliography{I_not_I}
\end{document}